\documentclass{article}

\def \TRtitle{Conic Multi-Task Classification}
\def \TRauthors{Cong Li, Michael Georgiopoulos and Georgios C. Anagnostopoulos}
\def \TRemails{congli@eecs.ucf.edu, michaelg@ucf.edu and georgio@fit.edu}
\def \TRkeywords{Multiple Kernel Learning, Multi-task Learning, Statistical Learning Theory, Generalization Bound, Multi-objective Optimization, Support Vector Machines}

\usepackage{options}
\usepackage{packages}
\usepackage{macros}

\begin{document} 

\maketitle

\ifMakeReviewDraft
	\linenumbers
\fi

\vskip 0.5in
\noindent
{\bf Keywords:} \TRkeywords

\begin{abstract} 

Traditionally, Multi-task Learning (MTL) models optimize the average of task-related objective functions, which is an intuitive approach and which we will be referring to as Average MTL. However, a more general framework, referred to as Conic MTL, can be formulated by considering conic combinations of the objective functions instead; in this framework, Average MTL arises as a special case, when all combination coefficients equal $1$. Although the advantage of Conic MTL over Average MTL has been shown experimentally in previous works, no theoretical justification has been provided to date. In this paper, we derive a generalization bound for the Conic MTL method, and demonstrate that the tightest bound is not necessarily achieved, when all combination coefficients equal $1$; hence, Average MTL may not always be the optimal choice, and it is important to consider Conic MTL. As a byproduct of the generalization bound, it also theoretically explains the good experimental results of previous relevant works. Finally, we propose a new Conic MTL model, whose conic combination coefficients minimize the generalization bound, instead of choosing them heuristically as has been done in previous methods. The rationale and advantage of our model is demonstrated and verified via a series of experiments by comparing with several other methods.

\end{abstract}

\acresetall

\section{Introduction}
\label{sec:Introduction}

\ac{MTL} has been an active research field for over a decade, since its inception in \cite{Caruana1997}. By training multiple tasks simultaneously with shared information, it is expected that the generalization performance of each task can be improved, compared to training each task separately. Previously, various \ac{MTL} schemes have been considered, many of which model the $t$-th task by a linear function with weight $\boldsymbol{w}_t, t = 1, \cdots T$, and assume a certain, underlying relationship between tasks. For example, the authors in \cite{Evgeniou2004} assumed all $\boldsymbol{w}_t$'s to be part of a cluster centered at $\bar{\boldsymbol{w}}$, the latter one being learned jointly with $\boldsymbol{w}_t$. This assumption was further extended to the case, where the weights $\boldsymbol{w}_t$'s can be grouped into different clusters instead of a single global cluster \cite{Zhong2012,Zhou2011a}. Furthermore, a widely held \ac{MTL} assumption is that tasks share a common, potentially sparse, feature representation, as done in  \cite{Obozinski2010,Jalali2010,Gong2013,Liu2009a,Fei2011,Argyriou2008,Kang2011}, to name a few. It is worth mentioning that many of these works allow features to be shared among only a subset of tasks, which are considered ``similar'' or ``related'' to each other, where the relevance between tasks is discovered during training. This approach reduces and, sometimes, completely avoids the effect of ``negative transfer'', \ie, knowledge transferred between irrelevant tasks, which leads to degraded generalization performance. Several other recent works that focused on the discovery of task relatedness include \cite{Zhang2012,Zhang2013,Romera2012,Pu2013}. Additionally, some kernel-based \ac{MTL} models assume that the data from all tasks are pre-processed by a (partially) common feature mapping, thus (partially) sharing the same kernel function; see \cite{Tang2009,Rakotomamonjy2011,Samek2011}, again, to name a few.

Most of these previous \ac{MTL} formulations consider the following classic setting: A set of training data $\{ \boldsymbol{x}_t^i, y_t^i \} \in \mathcal{X} \times \mathcal{Y}, i = 1, \cdots, N_t$ is provided for the $t$-th task ($t = 1, \cdots, T$), where $\mathcal{X}$, $\mathcal{Y}$ are the input and output spaces correspondingly. Each datum from the $t$-th task is assumed to be drawn from an underlying probability distribution $P_t(X_t, Y_t)$, where $X_t$ and $Y_t$ are random variables in the input and output space respectively. Then, a \ac{MTL} problem is formulated as follows

\begin{equation}
	\min_{\boldsymbol{w} \in \Omega(\boldsymbol{w})} \; \sum_{t=1}^T f(\boldsymbol{w}_t, \boldsymbol{x}_t, \boldsymbol{y}_t)
\label{eq:classic_mtl}
\end{equation}

\noindent
where $\boldsymbol{w} \triangleq (\boldsymbol{w}_1, \cdots, \boldsymbol{w}_T)$ is the collection of all $\boldsymbol{w}_t$'s, and, similarly, $\boldsymbol{x}_t \triangleq (\boldsymbol{x}_t^1, \cdots, \boldsymbol{x}_t^{N_t})$, $\boldsymbol{y}_t \triangleq (y_t^1, \cdots, y_t^{N_t})$. $f$ is a function common to all tasks. It is important to observe that, without the constraint $\boldsymbol{w} \in \Omega(\boldsymbol{w})$, \pref{eq:classic_mtl} degrades to $T$ independent learning problems. Therefore, in most scenarios, the set $\Omega(\boldsymbol{w})$ is designed to capture the inter-task relationships. For example, in \cite{Tang2009}, the model combines \ac{MTL} with \ac{MKL}, which is formulated as follows

\begin{equation}
\begin{aligned}
	& f(\boldsymbol{w}_t, \boldsymbol{x}_t, \boldsymbol{y}_t)  \triangleq \frac{1}{2} \| \boldsymbol{w}_t \|^2 + C \sum_{i=1}^{N_t} l(\boldsymbol{w}_t, \phi_t(\boldsymbol{x}_t^i), \boldsymbol{y}_t^i)\\
	& \Omega(\boldsymbol{w})  \triangleq \{\boldsymbol{w} = (\boldsymbol{w}_1, \cdots, \boldsymbol{w}_T): \boldsymbol{w}_t \in \mathcal{H}_{\boldsymbol{\theta},\boldsymbol{\gamma}_t}, \boldsymbol{\theta} \in \Omega(\boldsymbol{\theta}), \boldsymbol{\gamma} \in \Omega(\boldsymbol{\gamma})\}
\end{aligned}
\label{eq:tang_mtl}
\end{equation}

\noindent
Here, $l$ is a specified loss function, $\phi_t : \mathcal{X} \rightarrow \mathcal{H}_{\boldsymbol{\theta},\boldsymbol{\gamma}_t}$ is the feature mapping for the $t$-th task, $\mathcal{H}_{\boldsymbol{\theta},\boldsymbol{\gamma}_t}$ is the \ac{RKHS} with reproducing kernel function $k_t \triangleq \sum_{m=1}^M (\theta_m + \gamma_t^m) k_m$, where $k_m: \mathcal{X} \times \mathcal{X} \rightarrow \mathbb{R}, m = 1,\cdots,M$ are pre-selected kernel functions. $\| \boldsymbol{w}_t \| \triangleq \sqrt{\langle \boldsymbol{w}_t, \boldsymbol{w}_t \rangle}$ is the norm defined in $\mathcal{H}_{\boldsymbol{\theta},\boldsymbol{\gamma}_t}$. Also, $\Omega(\boldsymbol{\theta})$ is the feasible set of $\boldsymbol{\theta} \triangleq (\theta_1, \cdots, \theta_M)$, and, similarly, $\Omega(\boldsymbol{\gamma})$ is the feasible set of $\boldsymbol{\gamma} \triangleq (\boldsymbol{\gamma}_1, \cdots, \boldsymbol{\gamma}_T)$. It is not hard to see that, in this setting, $\Omega(\boldsymbol{w})$ is designed such that all tasks partially share the same kernel function in a \ac{MKL} manner, parameterized by the common coefficient $\boldsymbol{\theta}$ and task-specific coefficient $\boldsymbol{\gamma}_t, t = 1,\cdots,T$.

Another example, Sparse \ac{MTL} \cite{Rakotomamonjy2011}, has the following formulation:

\begin{equation}
\begin{aligned}
	& f(\boldsymbol{w}_t, \boldsymbol{x}_t, \boldsymbol{y}_t)  \triangleq \sum_{i=1}^{N_t} l(\boldsymbol{w}_t, \phi_t(\boldsymbol{x}_t^i), \boldsymbol{y}_t^i)\\ 
	& \Omega(\boldsymbol{w})  \triangleq \{\boldsymbol{w} = (\boldsymbol{w}_1, \cdots, \boldsymbol{w}_T): \boldsymbol{w}_t \triangleq (\boldsymbol{w}_t^1, \cdots, \boldsymbol{w}_t^M), \sum_{m=1}^M(\sum_{t=1}^T \| \boldsymbol{w}_t^m \|^q)^{p/q} \leq R\}
\label{eq:sparse_mtl}
\end{aligned}
\end{equation}

\noindent
where $\boldsymbol{w}_t^m \in \mathcal{H}_m, \forall m = 1, \cdots, M, t = 1, \cdots, T$, $\boldsymbol{w}_t \in \mathcal{H}_1 \times \cdots \times \mathcal{H}_M$, $0 < p \leq 1$, $1 \leq q \leq 2$. Note that although the original Sparse \ac{MTL} is formulated as follows

\begin{equation}
	\min_{\boldsymbol{w}} \; \sum_{m=1}^M(\sum_{t=1}^T \| \boldsymbol{w}_t^m \|^q)^{p/q} + C \sum_{t=1}^{T}\sum_{i=1}^{N_t} l(\boldsymbol{w}_t, \phi_t(\boldsymbol{x}_t^i), \boldsymbol{y}_t^i)
\label{eq:sparse_mtl_original}
\end{equation}

\noindent
due to the first part of Proposition $12$ in \cite{Kloft2011}, which we restate as \propref{prop:kloft2011} below\footnote{Note that the difference between \propref{prop:kloft2011} here and Proposition 12 in \cite{Kloft2011} is that, \propref{prop:kloft2011} does not require convexity of $f$, $g$ and $\mathcal{D}$; these are requirements necessary for the second part of Proposition 12 in \cite{Kloft2011}, which we do not utilize here.}, it is obvious that, for any $C > 0$, there exists a $R > 0$, such that \pref{eq:classic_mtl} 
and \pref{eq:sparse_mtl_original} are equivalent.

\begin{proposition}
\label{prop:kloft2011}
Let $\mathcal{D} \subseteq \mathcal{X}$, and let $f, g : \mathcal{D} \mapsto \mathbb{R}$ be two functions. For any $\sigma > 0$, there must exist a $\tau > 0$, such that the following two problems are equivalent

\begin{equation}
\min_{x \in \mathcal{D}} \; f(x) + \sigma g(x)
\label{eq:equivalent_1}
\end{equation}

\begin{equation}
\min_{x \in \mathcal{D}, g(x) \leq \tau} \; f(x)
\label{eq:equivalent_2}
\end{equation}
\end{proposition}

\noindent
The formulation given in \pref{eq:classic_mtl}, which we refer to as \textit{Average \ac{MTL}}, is intuitively appealing: It is reasonable to expect the average generalization performance of the $T$ tasks to be improved, by optimizing the average of the $T$ objective functions. However, as argued in \cite{Li2013}, solving \pref{eq:classic_mtl} yields only a particular solution on the Pareto Front of the following \ac{MOO} problem

\begin{equation}
\min_{\boldsymbol{w} \in \Omega(\boldsymbol{w})} \; \boldsymbol{f}(\boldsymbol{w}, \boldsymbol{x}, \boldsymbol{y})
\label{eq:moo}
\end{equation}

\noindent
where $\boldsymbol{f}(\boldsymbol{w}, \boldsymbol{x}, \boldsymbol{y}) \triangleq [f(\boldsymbol{w}_1, \boldsymbol{x}_1, \boldsymbol{y}_1), \cdots, f(\boldsymbol{w}_T, \boldsymbol{x}_T, \boldsymbol{y}_T)]'$. This is true, because scalarizing a \ac{MOO} problem by optimizing different conic combinations of the objective functions, leads to the discovery of solutions that correspond to points on the convex part of the problem's Pareto Front \cite[p. 178]{Boyd2004}. In other words, by conically scalarizing \pref{eq:moo} using different $\boldsymbol{\lambda} \triangleq [\lambda_1, \cdots, \lambda_T]'$, $\lambda_t > 0, \forall t=1,\cdots,T$, the optimization problem

\begin{equation}
\min_{\boldsymbol{w} \in \Omega(\boldsymbol{w})} \; \sum_{t=1}^T \lambda_t f(\boldsymbol{w}_t, \boldsymbol{x}_t, \boldsymbol{y}_t)
\label{eq:conic_comb}
\end{equation}

\noindent
yields different points on the Pareto Front of \pref{eq:moo}. Therefore, there is little reason to believe that the solution of \pref{eq:conic_comb} for the special case of $\lambda_t = 1, \forall t=1,\cdots,T$, \ie, the Average \ac{MTL}'s solution, is the best achievable. In fact, there might be other points on the Pareto Front that result in better generalization performance for each task, hence, yielding better average performance of the $T$ tasks.  Therefore, instead of solving \pref{eq:classic_mtl}, one can accomplish this by optimizing \pref{eq:conic_comb}.

A previous work along these lines was performed in \cite{Li2013}. The authors considered the following \ac{MTL} formulation, named \textit{Pareto-Path \ac{MTL}}

\begin{equation}
\min_{\boldsymbol{w} \in \Omega(\boldsymbol{w})} \; [\sum_{t=1}^T (f(\boldsymbol{w}_t, \boldsymbol{x}_t, \boldsymbol{y}_t))^p]^{1/p}
\label{eq:pareto_mtl}
\end{equation}

\noindent
which, assuming all objective functions are positive, minimizes the $L_p$-norm of the objectives when $p \geq 1$, and the $L_p$-pseudo-norm when $0 < p < 1$. It was proven that, for any $p > 0$, \pref{eq:pareto_mtl} is equivalent to \pref{eq:conic_comb} with

\begin{equation}
\lambda_t = \begin{cases}
\frac{f(\boldsymbol{w}_t, \boldsymbol{x}_t, \boldsymbol{y}_t)^{p-1}}{\sum_{t=1}^T (f(\boldsymbol{w}_t, \boldsymbol{x}_t, \boldsymbol{y}_t))^p} & \text{ if } p > 1 \\ 
1 & \text{ if } p = 1 \\ 
\frac{\sum_{t=1}^T (f(\boldsymbol{w}_t, \boldsymbol{x}_t, \boldsymbol{y}_t))^{\frac{1-p}{p}}}{f(\boldsymbol{w}_t, \boldsymbol{x}_t, \boldsymbol{y}_t)^{1-p}} & \text{ if } 0 < p < 1
\end{cases}, \forall t = 1, \cdots, T
\label{eq:pareto_mtl_lambda}
\end{equation}

\noindent
Thus by varying $p > 0$, the solutions of \pref{eq:pareto_mtl} trace a path on the Pareto Front of \pref{eq:moo}. While Average MTL is equivalent to \pref{eq:pareto_mtl}, when $p=1$, it was demonstrated that the experimental results are usually better when $p < 1$, compared to $p = 1$, in a \ac{SVM}-based \ac{MKL} setting. Regardless of the close correlation of the superior obtained results to our previous argument, the authors did not provide a rigorous basis of the advantage of considering an objective function other than the average of the $T$ task objectives. Therefore, use of the $L_p$-(pseudo-)norm in the paper's objective function remains so far largely a heuristic element of their approach.

In light of the just-mentioned potential drawbacks of Average MTL and the lack of supporting theory in the case of Pareto-Path MTL, in this paper, we analytically justify why it is worth considering \pref{eq:conic_comb}, which we refer to as \textit{Conic MTL}, and why it is advantageous. Specifically, a major contribution of this paper is the derivation of a generalization bound for Conic MTL, which illustrates that, indeed, the tightest bound is not necessarily achieved, when all $\lambda_t$'s equal to $1$. Therefore, it answers the previous question, and justifies the importance of considering Conic MTL. 
Also, as a byproduct of the generalization bound, in \sref{sec:Generalization_bound}, we theoretically show the benefit of Pareto-Path MTL: the generalization bound of \pref{eq:pareto_mtl} is usually tighter when $p < 1$, compared to the case, when $p = 1$. Therefore, it explains Pareto-Path MTL's superiority over Average MTL. 

Regarding Conic MTL, a natural question is how to choose the coefficients $\lambda_t$'s. Instead of setting them heuristically, such as what Pareto-Path MTL does, we propose a new Conic MTL model that learns the $\lambda_t$'s by minimizing the generalization bound. It ensures that our new model achieves the tightest generalization bound compared to any other settings of the $\lambda_t$ values and, potentially, leads to superior performance. The new model is described in \sref{sec:Model} and experimentally evaluated in \sref{sec:Experiment}. The experimental results verified our theoretical conclusions: Conic MTL can indeed outperform Average MTL and Pareto-Path MTL in many scenarios and, therefore, learning the coefficients $\lambda_t$'s by minimizing the generalization bound is reasonable and advantageous. Finally, we summarize our work in \sref{sec:Conclusion}.

In the sequel, we'll be using the following notational conventions: vector and matrices are denoted in boldface. Vectors are assumed to be columns vectors. If $\boldsymbol{v}$ is a vector, then $\boldsymbol{v}'$ denotes the transposition of $\boldsymbol{v}$. Vectors $\boldsymbol{0}$ and $\boldsymbol{1}$ are the all-zero and all-one vectors respectively. Also, $\succeq$, $\succ$, $\preceq$ and $\prec$ between vectors will stand for the component-wise $\geq$, $>$, $\leq$ and $<$ relations respectively. Similarly, for any $\boldsymbol{v}$, $\boldsymbol{v}^p$ represents the component-wise exponentiation of $\boldsymbol{v}$.

\section{Generalization Bound}
\label{sec:Generalization_bound}

Similar to previous theoretical analyses of \ac{MTL} methods \cite{Ando2005,Maurer2006,Maurer2006a,Kakade2012,Maurer2012,Pontil2013}, in this section, we derive the Rademacher complexity-based generalization bound for Conic MTL, \ie,  \pref{eq:conic_comb}. Specifically, we assume the following form of $f$ and $\Omega(\boldsymbol{w})$ for classification problems:

\begin{equation}
\begin{aligned}
	& f(\boldsymbol{w}_t, \boldsymbol{x}_t, \boldsymbol{y}_t)  \triangleq \frac{1}{2} \| \boldsymbol{w}_t \|^2 + C \sum_{i=1}^{N} l(y_t^i\langle \boldsymbol{w}_t, \phi(\boldsymbol{x}_t^i) \rangle)\\ 
	& \Omega(\boldsymbol{w})  \triangleq \{\boldsymbol{w} = (\boldsymbol{w}_1, \cdots, \boldsymbol{w}_T): \boldsymbol{w}_t \in \mathcal{H}_{\boldsymbol{\theta}}, \boldsymbol{\theta} \in \Omega(\boldsymbol{\theta}) \}
\end{aligned}
\label{eq:define_f}
\end{equation}

\noindent
where $l$ is the margin loss:

\begin{equation}
l(x) = \begin{cases}
0 & \text{ if } \rho \leq x \\ 
1-x/\rho & \text{ if } 0 \leq x \leq \rho \\ 
1 & \text{ if } x \leq 0
\end{cases}
\label{eq:margin_loss}
\end{equation}

\noindent
$\phi : \mathcal{X} \rightarrow \mathcal{H}_{\boldsymbol{\theta}}$ is the common feature mapping for all tasks. $\mathcal{H}_{\boldsymbol{\theta}}$ is the \ac{RKHS} defined by the kernel function $k \triangleq \sum_{m=1}^M \theta_m k_m$, where $k_m: \mathcal{X} \times \mathcal{X} \rightarrow \mathbb{R}, m = 1,\cdots,M$ are the pre-selected kernel functions. Furthermore, we assume the training data $\{ \boldsymbol{x}_t^i, y_t^i \} \in \mathcal{X} \times \mathcal{Y}, t=1,\cdots,T, i=1,\cdots,N$ are drawn from the probability distribution $P_t(X_t, Y_t)$, where $X_t$ and $Y_t$ are random variables in the input and output space respectively. Note that, here, we assumed all tasks have equal number of training data and share a common kernel function. These two assumptions were made to simplify notation and exposition, and they do not affect extending our results to a more general case, where an arbitrary number of training samples is available for each task and partially shared kernel functions are used; in the latter case, only relevant tasks may share the common kernel function, hence, reducing the effect of ``negative transfer''.

Substituting (\ref{eq:define_f}) into \pref{eq:conic_comb} and based on \propref{prop:kloft2011}, it is not hard to see that for any $C$ in \eref{eq:define_f}, there exist a $R > 0$ such that \pref{eq:conic_comb} is equivalent to the following problem

\begin{equation}
\begin{aligned}
\min_{\boldsymbol{w} \in \Omega(\boldsymbol{w})} \; & \sum_{t=1}^T \sum_{i=1}^{N_t} \lambda_t l(y_t^i\langle \boldsymbol{w}_t, \phi(\boldsymbol{x}_t^i) \rangle) \\
\textit{s.t.} & \sum_{t=1}^T \lambda_t \| \boldsymbol{w}_t \|^2 \leq R
\end{aligned}
\label{eq:conic_comb_equi}
\end{equation}

Obviously, solving \pref{eq:conic_comb_equi} is the process of choosing the $\boldsymbol{w}$ in the hypothesis space $\mathcal{F}_{\boldsymbol{\lambda}}$, such that the empirical loss, \ie, the objective function of \pref{eq:conic_comb_equi}, is minimized. The relevant hypothesis space is defined below:

\begin{equation}
\mathcal{F}_{\boldsymbol{\lambda}} \triangleq \{\boldsymbol{w} = (\boldsymbol{w}_1, \cdots, \boldsymbol{w}_T) : \sum_{t=1}^T \lambda_t \| \boldsymbol{w}_t \|^2 \leq R, \boldsymbol{w}_t \in \mathcal{H}_{\boldsymbol{\theta}}, \boldsymbol{\theta} \in \Omega(\boldsymbol{\theta}) \}
\label{eq:f_lambda}
\end{equation}

By defining the Conic MTL expected error $er(\boldsymbol{w})$ and empirical loss $\hat{er}_{\boldsymbol{\lambda}}(\boldsymbol{w})$ as follows

\begin{equation}
er(\boldsymbol{w}) = \frac{1}{T} \sum_{t=1}^T E[\boldsymbol{1}_{(-\infty, 0]}(Y_t\langle \boldsymbol{w}_t, \phi(X_t) \rangle)]
\label{eq:expected_error}
\end{equation}

\begin{equation}
\hat{er}_{\boldsymbol{\lambda}}(\boldsymbol{w}) = \frac{1}{TN} \sum_{t=1}^T \sum_{i=1}^{N} \lambda_t l(y_t^i\langle \boldsymbol{w}_t, \phi(\boldsymbol{x}_t^i) \rangle)
\label{eq:empirical_loss}
\end{equation}

\noindent
one of our major contribution is the following theorem, which gives the generalization bound of \pref{eq:conic_comb_equi} in the context of \ac{MKL}-based Conic MTL for any $\lambda_t \in (1, r_\lambda), \forall t=1, \cdots, T$, where $r_\lambda$ is a pre-specified upper-bound for the $\lambda_t$'s.

\begin{theorem}
\label{thm:bound_any_lambda}
For fixed $\rho > 0$, $r_{\lambda} \in \mathbb{N}$ with $r_{\lambda} > 1$, and for any $\boldsymbol{\lambda} = [\lambda_1, \cdots, \lambda_T]'$, $\lambda_t \in (1, r_{\lambda}), \forall t = 1, \cdots, T$, $\boldsymbol{w} \in \mathcal{F}_{\boldsymbol{\lambda}}$, $0 < \delta < 1$, the following generalization bound holds with probability at least $1-\delta$:

\begin{equation}
er(\boldsymbol{w}) \leq \hat{er}_{\boldsymbol{\lambda}}(\boldsymbol{w}) + \frac{\sqrt{2} r_{\lambda}}{\rho} R(\mathcal{F}_{\boldsymbol{\lambda}}) + \sqrt{\frac{9}{TN}\ln  \left( \frac{2 r_{\lambda}}{T} \sum_{t=1}^T \frac{1}{\lambda_t} \right) } + \sqrt{\frac{9\ln \frac{1}{\delta}}{2TN}}
\label{eq:bound_any_lambda}
\end{equation}

\noindent
where $R(\mathcal{F}_{\boldsymbol{\lambda}})$ is the empirical Rademacher complexity of the hypothesis space $\mathcal{F}_{\boldsymbol{\lambda}}$, which is defined as

\begin{equation}
R(\mathcal{F}_{\boldsymbol{\lambda}}) \triangleq \frac{2}{TN}E[\sup_{\boldsymbol{w} \in \mathcal{F}_{\boldsymbol{\lambda}}} \sum_{t=1}^T \sum_{i=1}^N \sigma_t^i \langle \boldsymbol{w}_t, \phi(\boldsymbol{x}_t^i) \rangle]
\label{eq:erc}
\end{equation}

\noindent
and the $\sigma_t^i$'s are i.i.d. Rademacher-distributed (\ie, $\mathrm{Bernoulli}(1/2)$-distributed random variables with sample space $\{-1, +1\}$).
\end{theorem}

Based on \thmref{thm:bound_any_lambda}, one is motivated to choose $\boldsymbol{\lambda}$ that minimizes the generalization bound, instead of heuristically selecting $\boldsymbol{\lambda}$ as in \eref{eq:pareto_mtl_lambda}, which was suggested in \cite{Li2013}. Indeed, doing so does not guarantee obtaining the tightest generalization bound.

However, prior to proposing our new Conic MTL model that minimizes the generalization bound, it is still of interest to theoretically analyze why Pareto-Path MTL, \ie, \pref{eq:pareto_mtl}, usually enjoys better generalization performance when $0 < p < 1$, rather than when $p = 1$, as described in \sref{sec:Introduction}. While the analysis is not given in \cite{Li2013}, fortunately, we can provide some insights of the good performance of the model, when $0 < p < 1$, by utilizing \thmref{thm:bound_any_lambda} and with the help of the following two theorems.

\begin{theorem}
\label{thm:monotone_erc}
For $\boldsymbol{\lambda} \succ \boldsymbol{0}$, the empirical Rademacher complexity $R(\mathcal{F}_{\boldsymbol{\lambda}})$ is monotonically decreasing with respect to each $\lambda_t, t = 1,\cdots,T$.
\end{theorem}

\begin{theorem}
\label{thm:monotone_lambda}
Assume $f(\boldsymbol{w}_t, \boldsymbol{x}_t, \boldsymbol{y}_t) > 0, \forall t = 1,\cdots,T$. For $\boldsymbol{\lambda}$ that is defined in \eref{eq:pareto_mtl_lambda}, when $0 < p < 1$, we have $\lambda_t > 1$ and $\lambda_t$ is monotonically decreasing with respect to $p$, $\forall t = 1, \cdots, T$.
\end{theorem}

Based on \eref{eq:pareto_mtl_lambda}, if $f(\boldsymbol{w}_t, \boldsymbol{x}_t, \boldsymbol{y}_t) > 0, \forall t = 1,\cdots,T$, there must exist a fixed $r_{\lambda} > 0$, such that $\lambda_t \in (1, r_{\lambda}),\forall t = 1, \cdots, T$. Therefore we can analyze the generalization bound of Pareto-Path MTL based on \thmref{thm:bound_any_lambda}, when $0 < p < 1$. Although \thmref{thm:bound_any_lambda} is not suitable for the case when $p=1$, we can approximate its bound by letting $p$ to be infinitely close to $1$.

The above two theorems indicate that the empirical Rademacher complexity for the hypothesis space of Pareto-Path MTL monotonically increases with respect to $p$, when $0 < p < 1$. Therefore, the second term in the generalization bound decreases as $p$ decreases. This is also true for the third term in the bound, based on \thmref{thm:monotone_lambda}. Thus, it is not a surprise that the generalization performance is usually better when $0 < p < 1$ than when $p = 1$, and it is reasonable to expect the performance to get improved when $p$ decreases. In fact, such a monotonicity is reported in the experiments of \cite{Li2013}: the classification accuracy is usually monotonically increasing, when $p$ decreases. It is worth mentioning that, although rarely observed, we may not have such monotonicity in performance, if the first term in the generalization bound, \ie, the empirical loss, grows quickly as $p$ decreases. However, the monotonic behavior of the generalization bound (except the empirical loss) is still sufficient for explaining the experimental results of \pref{eq:pareto_mtl}, which justifies the rationale of employing an arbitrarily weighted conic combination of objective functions instead of using the average of these functions.

Finally, we provide two theorems that not only are used in the proof of \thmref{thm:bound_any_lambda}, but also may be of interest on their own accord. Subsequently, in the next section, we describe our new \ac{MTL} model.

\begin{theorem}
\label{thm:monotone_erc_2}
Given $\boldsymbol{\gamma} \triangleq [\gamma_1, \cdots, \gamma_T]'$ with $\boldsymbol{\gamma} \succ \boldsymbol{0}$, define

\begin{equation}
R(\mathcal{F}_{\boldsymbol{\lambda}}, \boldsymbol{\gamma}) = \frac{2}{TN}E[\sup_{\boldsymbol{w} \in \mathcal{F}_{\boldsymbol{\lambda}}} \sum_{t=1}^T \sum_{i=1}^N \gamma_t \sigma_t^i \langle \boldsymbol{w}_t, \phi(\boldsymbol{x}_t^i) \rangle]
\label{eq:erc_gamma}
\end{equation}

\noindent
For fixed $\boldsymbol{\lambda} \succ \boldsymbol{0}$, $R(\mathcal{F}_{\boldsymbol{\lambda}}, \boldsymbol{\gamma})$ is monotonically increasing with respect to each $\gamma_t$.
\end{theorem}

\begin{theorem}
\label{thm:bound_fixed_lambda}
For fixed $r_{\lambda} \geq 1$, $\rho > 0$, $\boldsymbol{\lambda} = [\lambda_1, \cdots, \lambda_T]', \lambda_t \in [1, r_{\lambda}], \forall t = 1,\cdots,T$, and for any $\boldsymbol{w} \in \mathcal{F}_{\boldsymbol{\lambda}}$, $0 < \delta < 1$, the following generalization bound holds with probability at least $1-\delta$:

\begin{equation}
er(\boldsymbol{w}) \leq \hat{er}_{\boldsymbol{\lambda}}(\boldsymbol{w}) + \frac{r_{\lambda}}{\rho} R(\mathcal{F}_{\boldsymbol{\lambda}}) + \sqrt{\frac{9\ln \frac{1}{\delta}}{2TN}}
\label{eq:bound_fixed_lambda}
\end{equation}
\end{theorem}

Note that the difference between \thmref{thm:bound_fixed_lambda} and \thmref{thm:bound_any_lambda} is that, \thmref{thm:bound_any_lambda} is valid for \textit{any} $\lambda_t \in (1, r_{\lambda})$, while \thmref{thm:bound_fixed_lambda} is only valid for \textit{fixed} $\lambda_t \in [1, r_{\lambda}]$. While the bound given in \thmref{thm:bound_any_lambda} is more general, it is looser due to the additional third term in (\ref{eq:bound_any_lambda}) and due to the factor $\sqrt{2}$ multiplying the empirical Rademacher complexity.

\section{A New \ac{MTL} Model}
\label{sec:Model}

In this section, we propose our new \ac{MTL} model. Motivated by the generalization bound in \thmref{thm:bound_any_lambda}, our model is formulated to select $\boldsymbol{w}$ and $\boldsymbol{\lambda}$ by minimizing the bound

\begin{equation}
\hat{er}_{\boldsymbol{\lambda}}(\boldsymbol{w}) + \frac{\sqrt{2} r_{\lambda}}{\rho} R(\mathcal{F}_{\boldsymbol{\lambda}}) + \sqrt{\frac{9}{TN}\ln  \left( \frac{2 r_{\lambda}}{T} \sum_{t=1}^T \frac{1}{\lambda_t} \right) } + \sqrt{\frac{9\ln \frac{1}{\delta}}{2TN}}
\label{eq:bound_rhs}
\end{equation}

\noindent
instead of choosing the coefficients $\boldsymbol{\lambda}$ heuristically, such as via \eref{eq:pareto_mtl_lambda} in \cite{Li2013}. Note that the bound's last term does not depend on any model parameters, while the third term has only a minor effect on the bound, when $\lambda_t \in (1, r_{\lambda})$. Therefore, we omit these two terms, and propose the following model:

\begin{equation}
\begin{aligned}
\min_{\boldsymbol{w}, \boldsymbol{\lambda}} \; & \hat{er}_{\boldsymbol{\lambda}}(\boldsymbol{w}) + \frac{\sqrt{2} r_{\lambda}}{\rho} R(\mathcal{F}_{\boldsymbol{\lambda}}) \\ 
\textit{s.t.} \;& \boldsymbol{w} \in \mathcal{F}_{\boldsymbol{\lambda}},  \boldsymbol{1} \prec \boldsymbol{\lambda} \prec r_{\lambda}\boldsymbol{1}.
\end{aligned}
\label{eq:model_abstract_form}
\end{equation}

Furthermore, due to the complicated nature of $R(\mathcal{F}_{\boldsymbol{\lambda}})$, it is difficult to optimize \pref{eq:model_abstract_form} directly. Therefore, in the following theorem, we prove an upper bound for $R(\mathcal{F}_{\boldsymbol{\lambda}})$, which yields a simpler expression. We remind the readers that the hypothesis space $\mathcal{F}_{\boldsymbol{\lambda}}$ is defined as 

\begin{equation}
\mathcal{F}_{\boldsymbol{\lambda}} \triangleq \{\boldsymbol{w} = (\boldsymbol{w}_1, \cdots, \boldsymbol{w}_T) : \sum_{t=1}^T \lambda_t \| \boldsymbol{w}_t \|^2 \leq R, \boldsymbol{w}_t \in \mathcal{H}_{\boldsymbol{\theta}}, \boldsymbol{\theta} \in \Omega(\boldsymbol{\theta}) \}
\label{eq:f_lambda_again}
\end{equation}

\noindent
where $\mathcal{H}_{\boldsymbol{\theta}}$ is the \ac{RKHS} defined by the kernel function $k \triangleq \sum_{m=1}^M \theta_m k_m$.

\begin{theorem}
\label{thm:erc_upper_bound}
Given the hypothesis space $\mathcal{F}_{\boldsymbol{\lambda}}$, the empirical Rademacher complexity can be upper-bounded as follows:

\begin{equation}
R(\mathcal{F}_{\boldsymbol{\lambda}}) \leq \frac{2}{TN}\sqrt{\sum_{t=1}^{T} \frac{1}{\lambda_t}} \; E \left [\sqrt{ \sup_{\boldsymbol{w} \in \mathcal{F}_{\boldsymbol{1}}} \sum_{t=1}^T  \left( \sum_{i=1}^N \sigma_t^i \langle \boldsymbol{w}_t, \phi(\boldsymbol{x}_t^i) \rangle \right)^2} \; \right ]
\label{eq:erc_upper_bound}
\end{equation}

\noindent
where the feasible region of $\boldsymbol{w}$, \ie, $\mathcal{F}_{\boldsymbol{1}}$, is the same as $\mathcal{F}_{\boldsymbol{\lambda}}$ but with $\boldsymbol{\lambda} = \boldsymbol{1}$.
\end{theorem}

Note that, for a given $\Omega(\boldsymbol{\theta})$, the expectation term in (\ref{eq:erc_upper_bound}) is a constant. If we define

\begin{equation}
s \triangleq E \left [\sqrt{ \sup_{\boldsymbol{w} \in \mathcal{F}_{\boldsymbol{1}}} \sum_{t=1}^T \left( \sum_{i=1}^N \sigma_t^i \langle \boldsymbol{w}_t, \phi(\boldsymbol{x}_t^i) \rangle \right)^2} \; \right ]
\label{eq:rename_expectation}
\end{equation}

\noindent
we arrive at our proposed \ac{MTL} model:

\begin{equation}
\begin{aligned}
\min_{\boldsymbol{w}, \boldsymbol{\lambda}} \; & \sum_{t=1}^T \sum_{i=1}^{N} \lambda_t l(y_t^i\langle \boldsymbol{w}_t, \phi(\boldsymbol{x}_t^i) \rangle) + \frac{2\sqrt{2} s r_{\lambda}}{\rho} \sqrt{\sum_{t=1}^{T} \frac{1}{\lambda_t}} \\ 
\textit{s.t.} \;& \boldsymbol{w}_t \in \mathcal{H}_{\boldsymbol{\theta}}, \forall t = 1, \cdots, T \\
& \boldsymbol{\theta} \in \Omega(\boldsymbol{\theta}), \; \sum_{t=1}^{T} \lambda_t \| \boldsymbol{w}_t \|^2 \leq R,  \boldsymbol{1} \prec \boldsymbol{\lambda} \prec r_{\lambda}\boldsymbol{1}.
\end{aligned}
\label{eq:model}
\end{equation}

The next proposition provides an equivalent optimization problem, which is easier to solve.

\begin{proposition}
For any fixed $C > 0$, $s > 0$ and $r_{\lambda} > 0$, there exist $R > 0$ and $a > 0$ such that \pref{eq:model} and the following optimization problem are equivalent

\begin{equation}
\begin{aligned}
\min_{\boldsymbol{w}, \boldsymbol{\lambda}, \boldsymbol{\theta}} \; & \sum_{t=1}^{T} \lambda_t (\sum_{m=1}^{M}\frac{\| \boldsymbol{w}_t^m \|^2}{2 \theta_m} + C \sum_{i=1}^{N} \sum_{m=1}^{M} l(y_t^i\langle \boldsymbol{w}_t^m, \phi_m(\boldsymbol{x}_t^i) \rangle)) \\ 
\textit{s.t.} \; & \boldsymbol{w}_t^m \in \mathcal{H}_{m}, \forall t = 1, \cdots, T, m = 1, \cdots, M,\\
& \boldsymbol{\theta} \in \Omega(\boldsymbol{\theta}), \; \sum_{t=1}^{T} \frac{1}{\lambda_t} \leq a,  \boldsymbol{1} \prec \boldsymbol{\lambda} \prec r_{\lambda}\boldsymbol{1}. 
\end{aligned}
\label{eq:model_equivalent}
\end{equation}

\noindent
where $\mathcal{H}_m$ is the \ac{RKHS} defined by the kernel function $k_m$, and $\phi_m : \mathcal{X} \rightarrow \mathcal{H}_m$.
\label{prop:model_equivalent}
\end{proposition}

It is worth pointing out that, \pref{eq:model_equivalent} minimizes the generalization bound (\ref{eq:bound_rhs}) for \textit{any} $\Omega(\boldsymbol{\theta})$. 
A typical setting is to adapt the $L_p$-norm \ac{MKL} method by letting $\Omega(\boldsymbol{\theta}) \triangleq \{\boldsymbol{\theta} = [\theta_1, \cdots, \theta_M]' : \boldsymbol{\theta} \succeq \boldsymbol{0}, \| \boldsymbol{\theta} \|_p \leq 1 \}$, where $p \geq 1$. Alternatively, one may want to employ the \textit{optimal neighborhood kernel} method \cite{Liu2009} by letting $\Omega(\boldsymbol{\theta}) \triangleq \{\boldsymbol{\theta} = [\theta_1, \cdots, \theta_M]' : \sum_{t=1}^{T}\|
\boldsymbol{K}_t - \hat{\boldsymbol{K}_t} \|_F \leq R_k, \boldsymbol{K}_t \triangleq \sum_{m=1}^{M} \theta_m \boldsymbol{K}_t^m\}$, where $\boldsymbol{K}_t^m \in \mathbb{R}^{N \times N}$ is the kernel matrix whose $(i, j)$-th element is calculated as $k_m(\boldsymbol{x}_t^i, \boldsymbol{x}_t^j)$, and $\hat{\boldsymbol{K}_t}$'s are the kernel matrices evaluated by a pre-defined kernel function on the training data of the $t$-th task.

By assuming $\Omega(\boldsymbol{\theta})$ to be a convex set and electing the loss function $l$ to be convex in the model parameters (such as the hinge loss function), \pref{eq:model_equivalent} is jointly convex with respect to both $\boldsymbol{w}$ and $\boldsymbol{\theta}$. Also, it is separately convex with respect to $\boldsymbol{\lambda}$. Therefore, it is straightforward to employ a block-coordinate descent method to optimize \pref{eq:model_equivalent}. Finally, it is worth mentioning that, by choosing to employ the hinge loss function, the generalization bound in \thmref{thm:bound_any_lambda} still holds, since the hinge loss upper-bounds the margin loss for $\rho = 1$. Therefore, our model still minimizes the generalization bound.

\subsection{Incorporating \texorpdfstring{$L_p$-norm}{Lp-norm} \ac{MKL}}
\label{sec:mkl}

In this paper, we specifically consider endowing our \ac{MTL} model with $L_p$-norm \ac{MKL}, since it can be better analyzed theoretically, is usually easy to optimize and, often, yields good performance outcomes.

Although the upper bound in \thmref{thm:erc_upper_bound} is suitable for any $\Omega(\boldsymbol{\theta})$, it might be loose due to its generality. Another issue is that the expectation present in the bound is still hard to calculate. Therefore, as we consider $L_p$-norm \ac{MKL}, it is of interest to derive a bound specifically for it, which is easier to calculate and is potentially tighter.

\begin{theorem}
\label{thm:erc_upper_bound_mkl}
Let $\Omega(\boldsymbol{\theta}) \triangleq \{\boldsymbol{\theta} = [\theta_1, \cdots, \theta_M]' : \boldsymbol{\theta} \succeq \boldsymbol{0}, \| \boldsymbol{\theta} \|_p \leq 1 \}$, $p \geq 1$, and $\boldsymbol{K}_t^m \in \mathbb{R}^{N \times N}, t = 1,\cdots,T, m = 1,\cdots,M$ be the kernel matrix, whose $(i, j)$-th element is defined as $k_m(\boldsymbol{x}_t^i, \boldsymbol{x}_t^j)$. Also, define $\boldsymbol{v}_t \triangleq [tr(\boldsymbol{K}_t^1), \cdots, tr(\boldsymbol{K}_t^M)]' \in \mathbb{R}^M$. Then, we have

\begin{equation}
R(\mathcal{F}_{\boldsymbol{\lambda}}) \leq \frac{2\sqrt{2Rp^*}}{TN}  \sqrt{\sum_{t=1}^T \frac{1}{\lambda_t} \| \boldsymbol{v}_t \|_{p^*}}
\label{eq:erc_upper_bound_mkl}
\end{equation}

\noindent
where $p^* \triangleq \frac{p}{p-1}$.
\end{theorem}

Following a similar procedure to formulating our general model \pref{eq:model_equivalent}, we arrive at the following $L_p$-norm \ac{MKL}-based \ac{MTL} problem

\begin{equation}
\begin{aligned}
\min_{\boldsymbol{w}, \boldsymbol{\lambda}, \boldsymbol{\theta}} \; & \sum_{t=1}^{T} \lambda_t (\sum_{m=1}^{M}\frac{\| \boldsymbol{w}_t^m \|^2}{2 \theta_m} + C \sum_{i=1}^{N} \sum_{m=1}^{M} l(y_t^i\langle \boldsymbol{w}_t^m, \phi(\boldsymbol{x}_t^i) \rangle)) \\ 
\textit{s.t.} \; & \boldsymbol{w}_t^m \in \mathcal{H}_{m}, \forall t = 1, \cdots, T, m = 1, \cdots, M,\\
& \boldsymbol{\theta} \succeq \boldsymbol{0}, \| \boldsymbol{\theta} \|_p \leq 1, \\
& \sum_{t=1}^{T} \frac{\| \boldsymbol{v}_t \|_{p^*}}{\lambda_t} \leq a,  \boldsymbol{1} \prec \boldsymbol{\lambda} \prec r_{\lambda}\boldsymbol{1}.  
\end{aligned}
\label{eq:model_mkl}
\end{equation}

\noindent
which, based on (\ref{eq:bound_rhs}) and (\ref{eq:erc_upper_bound_mkl}), minimizes the generalization bound. Note that, due to the bound that is specifically derived for $L_p$-norm \ac{MKL}, the constraint $\sum_{t=1}^{T} \frac{1}{\lambda_t} \leq a$ in \pref{eq:model_equivalent} is changed to $\sum_{t=1}^{T} \frac{\| \boldsymbol{v}_t \|_{p^*}}{\lambda_t} \leq a$ in the previous problem. However, when all kernel matrices $\boldsymbol{K}_t^m$'s have the same trace (as is the case, when all kernel functions are normalized, such that $k_m(\boldsymbol{x}, \boldsymbol{x}) = 1, \forall m = 1, \cdots, M, \boldsymbol{x} \in \mathcal{X}$), for a given $p \geq 1$, $\| \boldsymbol{v}_t \|_{p^*}$ has the same value for all $t = 1, \cdots, T$. In this case, \pref{eq:model_mkl} is equivalent to \pref{eq:model_equivalent}.

\section{Experiments}
\label{sec:Experiment}

In this section, we conduct a series of experiments with several data sets, in order to show the merit of our proposed \ac{MTL} model by comparing it to a few other related methods.

\subsection{Experimental Settings}
\label{sec:Experimental_setting}
In our experiments, we specifically evaluate the $L_p$-norm \ac{MKL}-based \ac{MTL} model, \ie, \pref{eq:model_mkl}, on classification problems using the hinge loss function. To solve \pref{eq:model_mkl}, we employed a block-coordinate descent algorithm, which optimizes each of the three variables $\boldsymbol{w}$, $\boldsymbol{\lambda}$ and $\boldsymbol{\theta}$ in succession by holding the remaining two variables fixed. Specifically, in each iteration, three optimization problems are solved. First, for fixed $\boldsymbol{\lambda}$ and $\boldsymbol{\theta}$, the optimization with respect to $\boldsymbol{w}$ can be split into $T$ independent \ac{SVM} problems, which are solved via \texttt{LIBSVM} \cite{CC01a}. Next, for fixed $\boldsymbol{w}$ and $\boldsymbol{\theta}$, the optimization with respect to $\boldsymbol{\lambda}$ is convex and is solved using \texttt{CVX} \cite{Grant2008}\cite{cvx}. Finally, minimizing with respect to $\boldsymbol{\theta}$, while $\boldsymbol{w}$ and $\boldsymbol{\lambda}$ are held fixed, has a closed-form solution:

\begin{equation}
\boldsymbol{\theta}^* =\left (\frac{\boldsymbol{v}}{\| \boldsymbol{v} \|_{\frac{p}{p+1}}} \right )^{\frac{1}{p+1}}
\label{eq:theta}
\end{equation}

\noindent
where $\boldsymbol{v} \triangleq [v_1, \cdots, v_M]'$ and $v_m \triangleq \sum_{t=1}^T \| \boldsymbol{w}_t^m \|, \forall m = 1,\cdots,M$. Although more efficient algorithms may exist, we opted to use this simple and easy-to-implement algorithm, since the optimization strategy is not the focus of our paper\footnote{Our \texttt{MATLAB} implementation is located at \url{http://github.com/congliucf/ECML2014}}.

For all experiments, $11$ kernels were selected for use: a Linear kernel, a $2^{nd}$-order Polynomial kernel and Gaussian kernels with spread parameter values $\left \{ 2^{-7}, 2^{-5}, 2^{-3}, 2^{-1}, 2^{0}, 2^{1}, 2^{3}, 2^{5}, 2^{7} \right \}$. Parameters $C$, $p$ and $a$ were selected via cross-validation. Our model is evaluated on $6$ data sets: $2$ real-world data sets from the UCI repository \cite{Frank2010}, $2$ handwritten digits data sets, and $2$ multi-task data sets, which we detail below.

The Wall-Following Robot Navigation (\textit{Robot}) and Vehicle Silhouettes (\textit{Vehicle}) data sets were obtained from the UCI repository. The \textit{Robot} data, consisting of $4$ features per sample, describe the position of the robot, while it navigates through a room following the wall in a clockwise direction. Each sample is to be classified according to one of the following four classes: ``Move-Forward'', ``Slight-Right-Turn'', ``Sharp-Right-Turn'' and ``Slight-Left-Turn''. On the other hand, the \textit{Vehicle} data set is a collection of $18$-dimensional feature vectors extracted from images. Each datum should be classified into one of four classes: ``4 Opel'', ``SAAB'', ``Bus'' and ``Van''.

The two handwritten digit data sets, namely \textit{MNIST}\footnote{Available at: \url{http://yann.lecun.com/exdb/mnist/}} and \textit{USPS}\footnote{Available at: \url{http://www.cs.nyu.edu/~roweis/data.html}}, consist of grayscale images of handwritten digits from $0$ to $9$ with $784$ and $256$ features respectively. Each datum is labeled as one of ten classes, each of which represents a single digit. For these four multi-class data sets, an equal number of samples from each class were chosen for training. Also, we approached these multi-class problems as \ac{MTL} problems using a one-vs.-one strategy and the averaged classification accuracy is calculated for each data set.

The last two data sets, namely \textit{Letter}\footnote{Available at: \url{http://multitask.cs.berkeley.edu/}} and \textit{Landmine}\footnote{Available at: \url{http://people.ee.duke.edu/~lcarin/LandmineData.zip}}, correspond to pure multi-task problems. Specifically, the \textit{Letter} data set involves $8$ tasks: ``C'' vs.\ ``E'', ``G'' vs.\ ``Y'', ``M'' vs.\ ``N'', ``A'' vs.\ ``G'', ``I'' vs.\ ``J'', ``A'' vs.\ ``O'', ``F'' vs.\ ``T'' and ``H'' vs.\ ``N''. Each letter is represented by a $8 \times 16$ pixel image, which forms a $128$-dimensional feature vector. The goal for this problem is to correctly recognize the letters in each task. On the other hand, the \textit{Landmine} data set consists of $29$ binary classification tasks. Each datum is a $9$-dimensional feature vector extracted from radar images that capture a single region of landmine fields. The goal for each task is to detect landmines in specific regions. For the experiments involving these two data sets, we re-sampled the data such that, for each task, the two classes contain equal number of samples.

In all our experiments, we considered training set sizes of $10\%$, $20\%$ and $50\%$ of the original data set. As an exception, for the \textit{Landmine} data set, we did not use the $10\%$ of the original set for training due to its small size; instead, we used $20\%$, $30\%$ and $50\%$.

We compared our method with five different \ac{MT-MKL} methods. The first one is Pareto-Path MTL, \ie, \pref{eq:pareto_mtl}, which was originally proposed in \cite{Li2013}. One can expect our new method to outperform it in most cases, since our method selects $\boldsymbol{\lambda}$ by minimizing the generalization bound, while Pareto-Path MTL selects its value heuristically via \eref{eq:pareto_mtl_lambda}. The second method we compared with is the $L_p$-norm \ac{MKL}-based Average MTL, which is the same as our method for $\boldsymbol{\lambda} = \boldsymbol{1}$. As we argued earlier in the introduction, minimizing the averaged objective does not necessarily guarantee the best generalization performance. By comparing with Average MTL, we expect to verify our claim experimentally. Moreover, we compared with two other popular \ac{MT-MKL} methods, namely Tang's Method \cite{Tang2009} and Sparse MTL \cite{Rakotomamonjy2011}. These two methods were outlined in \sref{sec:Introduction}. Finally, we considered the baseline approach, which trains each task individually via a traditional single-task $L_p$-norm \ac{MKL} strategy.

\subsection{Experimental Results}
\label{sec:Experimental_results}

\tref{tab:result} provides the obtained experimental results  based on the settings that were described in the previous sub-section. More specifically, in \tref{tab:result}, we report the average classification accuracy of $20$ runs over a randomly sampled training set. Moreover, the best performance among the $6$ competing methods is highlighted in boldface. To test the statistical significance of the differences between our method and the $5$ other methods, we employed a t-test to compare mean accuracies using a significance level of $\alpha = 0.05$. In the table, underlined numbers indicate the results that are statistically significantly worse than the ones produced by our method.

\begin{table}[ht]
\begin{center}
\caption{Comparison of Multi-task Classification Accuracy between Our Method and Five Other Methods. Averaged performances of $20$ runs over randomly sampled training set are reported.}
\label{tab:result}
\begin{tabular}{l c c c c c c}
\toprule
Robot & Our Method & Pareto & Average & Tang & Sparse & Baseline \\
\midrule
$10\%$ & \textbf{95.83} & \underline{95.07} & \underline{95.16} & \underline{93.93} & \underline{94.69} & 95.54 \\
$20\%$ & \textbf{97.11} & \underline{96.11} & \underline{95.90} & \underline{96.36} & \underline{96.56} & \underline{95.75} \\
$50\%$ & \textbf{98.41} & \underline{96.80} & \underline{96.59} & \underline{97.21} & \underline{98.09} & \underline{96.31} \\
\midrule
Vehicle & Our Method & Pareto & Average & Tang & Sparse & Baseline \\
\midrule
$10\%$ & \textbf{80.10} & 80.05 & 79.77 & \underline{78.47} & \underline{79.28} & \underline{78.01} \\
$20\%$ & 84.69 & \textbf{85.33} & 85.22 & \underline{83.98} & 84.44 & 84.37 \\
$50\%$ & \textbf{89.90} & \underline{88.04} & \underline{87.93} & \underline{88.13} & \underline{88.57} & \underline{87.64} \\
\midrule
Letter & Our Method & Pareto & Average & Tang & Sparse & Baseline \\
\midrule
$10\%$ & 83.00 & \textbf{83.95} & \underline{81.45} & \underline{80.86} & 83.00 & \underline{81.33} \\
$20\%$ & 87.13 & \textbf{87.51} & \underline{86.42} & \underline{82.95} & 87.09 & \underline{86.39} \\
$50\%$ & 90.47 & \textbf{90.61} & 90.01 & \underline{84.87} & 90.65 & \underline{89.80} \\
\midrule
Landmine & Our Method & Pareto & Average & Tang & Sparse & Baseline \\
\midrule
$20\%$ & \textbf{70.18} & \underline{69.59} & \underline{67.24} & \underline{66.60} & \underline{58.89} & \underline{66.64} \\
$30\%$ & \textbf{74.52} & 74.15 & \underline{71.62} & \underline{70.89} & \underline{65.83} & \underline{71.14} \\
$50\%$ & \textbf{78.26} & 77.42 & \underline{76.96} & \underline{76.08} & \underline{75.82} & \underline{76.29} \\
\midrule
MNIST & Our Method & Pareto & Average & Tang & Sparse & Baseline \\
\midrule
$10\%$ & \textbf{93.59} & \underline{89.30} & \underline{88.81} & \underline{92.37} & 93.48 & \underline{88.71} \\
$20\%$ & \textbf{96.08} & \underline{95.02} & \underline{94.95} & 95.94 & 95.96 & \underline{94.81} \\
$50\%$ & 97.44 & \underline{96.92} & \underline{96.98} & 97.47 & \textbf{97.53} & \underline{97.04} \\
\midrule
USPS & Our Method & Pareto & Average & Tang & Sparse & Baseline \\
\midrule
$10\%$ & \textbf{94.61} & \underline{90.22} & \underline{90.11} & \underline{93.20} & 94.52 & \underline{89.02} \\
$20\%$ & \textbf{97.44} & \underline{96.26} & \underline{96.25} & 97.37 & 97.53 & \underline{96.17} \\
$50\%$ & \textbf{98.98} & \underline{98.51} & \underline{98.59} & 98.96 & \textbf{98.98} & \underline{98.49} \\
\bottomrule
\end{tabular}
\end{center}
\end{table}

When analyzing the results in \tref{tab:result}, first of all, we observe that the optimal result is almost always achieved by the two Conic MTL methods, namely our method and Pareto-Path MTL. This result not only shows the advantage of Conic MTL over Average MTL, but also demonstrates the benefit compared to other \ac{MTL} methods, such as Tang's MTL and Sparse MTL. Secondly, it is obvious that our method can usually achieve better result than Pareto-Path MTL; as a matter of fact, in many cases the advantage is statistically significant. This observation validates the underlying rationale of our method, which chooses the coefficient $\boldsymbol{\lambda}$ by minimizing the generalization bound instead of using \eref{eq:pareto_mtl_lambda}. Finally, when comparing our method against the five alternative methods, our results are statistically better most of the time, which further emphasizes the benefit of our method.

\section{Conclusions}
\label{sec:Conclusion}

In this paper, we considered the \ac{MTL} problem that minimizes the conic combination of objectives with coefficients $\boldsymbol{\lambda}$, which we refer to as Conic MTL. The traditional \ac{MTL} method, which minimizes the average of the task objectives (Average MTL), is only a special case of Conic MTL with $\boldsymbol{\lambda} = \boldsymbol{1}$. Intuitively, such a specific choice of $\boldsymbol{\lambda}$ should not necessarily lead to optimal generalization performance. 

This intuition motivated the derivation of a Rademacher complexity-based generalization bound for Conic MTL in a \ac{MKL}-based classification setting. The properties of the bound, as we have shown in \sref{sec:Generalization_bound}, indicate that the optimal choice of $\boldsymbol{\lambda}$ is indeed not necessarily equal to $\boldsymbol{1}$. Therefore, it is important to consider different values for $\boldsymbol{\lambda}$ for Conic MTL, which may yield tighter generalization bounds and, hence, better performance. As a byproduct, our analysis also explains the reported superiority of Pareto-Path MTL \cite{Li2013} over Average MTL.  

Moreover, we proposed a new Conic MTL model, which aims to directly minimize the derived generalization bound. Via a series of experiments on six widely utilized data sets, our new model demonstrated a statistically significant advantage over Pareto-Path MTL, Average MTL, and two other popular \ac{MT-MKL} methods.

\section*{Acknowledgments}
\label{sec:Acknowledgement}

Cong Li acknowledges support from National Science Foundation (NSF) grants No. 0806931 and No. 0963146. Furthermore, Michael Georgiopoulos acknowledges support from NSF grants No. 0963146, No. 1200566, and No. 1161228. Also, Georgios C. Anagnostopoulos acknowledges partial support from NSF grant No. 1263011. Note that any opinions, findings, and conclusions or recommendations expressed in this material are those of the authors and do not necessarily reflect the views of the NSF. Finally, the authors would like to thank the three anonymous reviewers, that reviewed this manuscript, for their constructive comments.
\bibliographystyle{plainurl}
\bibliography{references}

\newpage

\section*{Supplementary Material}
\label{sec:appendix}
In this supplementary material, we give the proofs of all the theoretical results. Before proving \thmref{thm:bound_any_lambda}, we first prove \thmref{thm:monotone_erc} to \thmref{thm:bound_fixed_lambda}, which are used in the proof of \thmref{thm:bound_any_lambda}.

\subsection{Proof to \thmref{thm:monotone_erc}}
\label{sec:proof_monotone_erc}
Let $\boldsymbol{\lambda}^{(1)} \triangleq [\lambda_1^{(1)}, \cdots, \lambda_T^{(1)}]'$, and $\boldsymbol{\lambda}^{(2)} \triangleq [\lambda_1^{(2)}, \cdots, \lambda_T^{(2)}]'$ with $\boldsymbol{\lambda^{(1)}} \succ \boldsymbol{0}$, $\boldsymbol{\lambda^{(2)}} \succ \boldsymbol{0}$. Suppose there exists a $t_0 \in \{1, \cdots, T\}$, such that

\begin{equation}
\begin{cases}
\lambda_t^{(1)} > \lambda_t^{(2)}& \text{ if } t = t_0 \\ 
\lambda_t^{(1)} = \lambda_t^{(2)} & \text{ if } t \neq t_0 
\end{cases}
\label{eq:proof_monotone_erc_1}
\end{equation}

\noindent
Then, for any $\boldsymbol{w} \in \mathcal{F}_{\boldsymbol{\lambda}^{(1)}}$, it must be true that $\boldsymbol{w} \in \mathcal{F}_{\boldsymbol{\lambda}^{(2)}}$. Therefore $\mathcal{F}_{\boldsymbol{\lambda}^{(1)}} \subseteq \mathcal{F}_{\boldsymbol{\lambda}^{(2)}}$, which means $R(\mathcal{F}_{\boldsymbol{\lambda}^{(1)}}) \leq R(\mathcal{F}_{\boldsymbol{\lambda}^{(2)}})$.

\subsection{Proof to \thmref{thm:monotone_lambda}}
\label{sec:proof_monotone_lambda}
First, it is obvious that $\boldsymbol{\lambda} \succ \boldsymbol{1}$ when $0 < p < 1$. So we only prove that $\lambda_t$ is monotonically decreasing with respect to $p$, $\forall t = 1,\cdots,T$. By letting $\zeta_t^* \triangleq 1/\lambda_t, \forall t = 1,\cdots,T$, where $\boldsymbol{\lambda}$ is given in \eref{eq:pareto_mtl_lambda}, it is proven in \cite{Li2013} that $\boldsymbol{\zeta}^* \triangleq [\zeta_1^*, \cdots, \zeta_T^*]'$ is the solution of the following problem:

\begin{equation}
\min_{\boldsymbol{\zeta} \in \bar{B}_q} \; \sum_{t=1}^T \frac{1}{\zeta_t} f(\boldsymbol{w}_t, \boldsymbol{x}_t, \boldsymbol{y}_t)
\label{eq:proof_monotone_lambda_1}
\end{equation}

\noindent
where $\bar{B}_q \triangleq \{\boldsymbol{\zeta} : \boldsymbol{\zeta} \succeq \boldsymbol{0}, (\sum_{t=1}^T \zeta_t^q)^{1/q} \leq 1\}$, and $q \triangleq \frac{p}{1-p}$. For any $q_1 > 0, q_2 > 0, q_1 \leq q_2$, let $\boldsymbol{\zeta}_1^*$ and $\boldsymbol{\zeta}_2^*$ be the solution of \pref{eq:proof_monotone_lambda_1} when $q = q_1$ and $q = q_2$, respectively. By observing that $\bar{B}_{q_1} \subseteq \bar{B}_{q_2}$, and, to minimize the objective function, each $\zeta_t$ is preferred to be as large as possible, we immediately know that $\boldsymbol{\zeta}_1^* \preceq \boldsymbol{\zeta}_2^*$, \ie, each $\zeta_t^*$ is monotonically increasing with respect to $q$. Finally, by observing that $q$ increases with $p$ and $\lambda_t = 1/\zeta_t^*$, we conclude that each $\lambda_t$ is monotonically decreasing with respect to $p$.

\subsection{Proof to \thmref{thm:monotone_erc_2}}
\label{sec:proof_monotone_erc_2}
For fixed $\boldsymbol{\lambda}$, let $\boldsymbol{v}_t \triangleq \gamma_t \boldsymbol{w}_t$, $\zeta_t \triangleq \lambda_t / \gamma_t^2, \forall t = 1,\cdots,T$, and substitute into \eref{eq:erc_gamma}. Then, we have that $R(\mathcal{F}_{\boldsymbol{\lambda}}, \boldsymbol{\gamma}) = R(\mathcal{F}_{\boldsymbol{\zeta}})$, where $\boldsymbol{\zeta} \triangleq [\zeta_1, \cdots, \zeta_T]'$. Based on \thmref{thm:monotone_erc}, we conclude that $R(\mathcal{F}_{\boldsymbol{\zeta}})$ is monotonically decreasing with respect to each $\zeta_t$, thus $R(\mathcal{F}_{\boldsymbol{\lambda}}, \boldsymbol{\gamma})$ is monotonically increasing with respect to each $\gamma_t$.

\subsection{Proof to \thmref{thm:bound_fixed_lambda}}
\label{sec:proof_bound_fixed_lambda}
We start with the definition of $er(\boldsymbol{w})$:

\begin{equation}
\begin{aligned}
er(\boldsymbol{w}) & = \frac{1}{T} \sum_{t=1}^T E[\boldsymbol{1}_{(-\infty, 0]}(Y_t\langle \boldsymbol{w}_t, \phi(X_t) \rangle)] \\
& = \frac{1}{T} \sum_{t=1}^T E[\boldsymbol{1}_{(-\infty, 0]}(\lambda_t Y_t\langle \boldsymbol{w}_t, \phi(X_t) \rangle)] \\
& \leq \frac{1}{T} \sum_{t=1}^T E[l(\lambda_t Y_t\langle \boldsymbol{w}_t, \phi(X_t) \rangle)] \\
\end{aligned}
\label{eq:proof_bound_fixed_lambda_1}
\end{equation}

\noindent
Based on Theorem 16 of \cite{Maurer2006a}, we have that, for any $\delta > 0$, with probability at least $1-\delta$:

\begin{equation}
\begin{aligned}
er(\boldsymbol{w}) & \leq \frac{1}{TN} \sum_{t=1}^T \sum_{i=1}^N l(\lambda_t y_t^i \langle \boldsymbol{w}_t, \phi(x_t^i) \rangle) \\
& + \frac{2}{TN}E[\sup_{\boldsymbol{w} \in \mathcal{F}_{\boldsymbol{\lambda}}} \sum_{t=1}^T \sum_{i=1}^N \sigma_t^i l(\lambda_t y_t^i \langle \boldsymbol{w}_t, \phi(x_t^i) \rangle)] + \sqrt{\frac{9 \ln \frac{1}{\delta}}{2TN}}
\end{aligned}
\label{eq:proof_bound_fixed_lambda_2}
\end{equation}

\noindent
First note that, when $\lambda_t \geq 1$, we have $l(\lambda_t y_t^i \langle \boldsymbol{w}_t, \phi(x_t^i) \rangle) \leq \lambda_t l(y_t^i \langle \boldsymbol{w}_t, \phi(x_t^i) \rangle)$, which gives

\begin{equation}
er(\boldsymbol{w}) \leq \hat{er}_{\boldsymbol{\lambda}}(\boldsymbol{w}) + \frac{2}{TN}E[\sup_{\boldsymbol{w} \in \mathcal{F}_{\boldsymbol{\lambda}}} \sum_{t=1}^T \sum_{i=1}^N \sigma_t^i l(\lambda_t y_t^i \langle \boldsymbol{w}_t, \phi(x_t^i) \rangle)] + \sqrt{\frac{9 \ln \frac{1}{\delta}}{2TN}}
\label{eq:proof_bound_fixed_lambda_3}
\end{equation}

\noindent
Second, based on the definition of the margin loss $l$, and Theorem 17 of \cite{Maurer2006a}, we have

\begin{equation}
\begin{aligned}
E[\sup_{\boldsymbol{w} \in \mathcal{F}_{\boldsymbol{\lambda}}} \sum_{t=1}^T \sum_{i=1}^N \sigma_t^i l(\lambda_t y_t^i \langle \boldsymbol{w}_t, \phi(x_t^i) \rangle)] & \leq \frac{1}{\rho} E[\sup_{\boldsymbol{w} \in \mathcal{F}_{\boldsymbol{\lambda}}} \sum_{t=1}^T \sum_{i=1}^N \sigma_t^i \lambda_t y_t^i \langle \boldsymbol{w}_t, \phi(x_t^i) \rangle] \\
& \leq \frac{r_{\lambda}}{\rho} E[\sup_{\boldsymbol{w} \in \mathcal{F}_{\boldsymbol{\lambda}}} \sum_{t=1}^T \sum_{i=1}^N \sigma_t^i  y_t^i \langle \boldsymbol{w}_t, \phi(x_t^i) \rangle]
\end{aligned}
\label{eq:proof_bound_fixed_lambda_4}
\end{equation}

\noindent
where the second inequality is due to \thmref{thm:monotone_erc_2}. The proof is completed by substituting (\ref{eq:proof_bound_fixed_lambda_4}) into (\ref{eq:proof_bound_fixed_lambda_3}).

\subsection{Proof to \thmref{thm:bound_any_lambda}}
\label{sec:proof_bound_any_lambda}
When $r_{\lambda} \in \mathbb{N}$, consider the sequence $\{ \epsilon_{\boldsymbol{k}} \}_{k_1, \cdots, k_T}$ and $\{ \boldsymbol{\lambda}_{\boldsymbol{k}} \}_{k_1, \cdots, k_T}$ with $k_t = 2, \cdots, r_{\lambda}$, where $\boldsymbol{\lambda}_{\boldsymbol{k}} \triangleq [\lambda_{k_1}, \cdots, \lambda_{k_T}]'$, $\lambda_{k_t} \triangleq \frac{r_{\lambda}}{k_t}$, $\epsilon_{\boldsymbol{k}} \triangleq \epsilon + \sqrt{\frac{9\ln \frac{\sum_{t=1}^{T} k_t}{T}}{TN}}$.

We first give the following inequality, which we will prove later:

\begin{equation}
P\{ \exists \boldsymbol{k} = [k_1, \cdots, k_T]' : er - \hat{er}_{\boldsymbol{\lambda}_{\boldsymbol{k}}} > \frac{r_{\lambda}}{\rho} R(\mathcal{F}_{\boldsymbol{\lambda}_{\boldsymbol{k}}}) + \epsilon_{\boldsymbol{k}} \} \leq exp \{- \frac{2TN \epsilon^2}{9}\}
\label{eq:proof_bound_any_lambda_1}
\end{equation}

\noindent
Then, note that $\forall \lambda_t \in (1, r_{\lambda})$, $\exists \; \hat{k}_t \in \mathbb{N}$ with $2\leq \hat{k}_t \leq r_{\lambda}$ such that $\lambda_t \in (\lambda_{\hat{k}_t}, \lambda_{\hat{k}_t - 1}]$. Therefore, for any $ \boldsymbol{1} \prec \boldsymbol{\lambda} \prec r_{\lambda} \boldsymbol{1}$, we must be able to find a $\hat{\boldsymbol{k}} \triangleq [\hat{k}_1, \cdots, \hat{k}_T]'$ such that $\boldsymbol{\lambda} \succ \boldsymbol{\lambda}_{\hat{\boldsymbol{k}}}$, and 

\begin{equation}
P\{ er - \hat{er}_{\boldsymbol{\lambda}_{\hat{\boldsymbol{k}}}} > \frac{r_{\lambda}}{\rho} R(\mathcal{F}_{\boldsymbol{\lambda}_{\hat{\boldsymbol{k}}}}) + \epsilon_{\hat{\boldsymbol{k}}} \} \leq exp \{- \frac{2TN \epsilon^2}{9}\}
\label{eq:proof_bound_any_lambda_2}
\end{equation}

\noindent
holds for any $\boldsymbol{w} \in \mathcal{F}_{\boldsymbol{\lambda}_{\hat{\boldsymbol{k}}}}$. Then we reach the conclusion that the following inequality 

\begin{equation}
P\{ er - \hat{er}_{\boldsymbol{\lambda}} > \frac{\sqrt{2} r_{\lambda}}{\rho} R(\mathcal{F}_{\boldsymbol{\lambda}}) + \sqrt{\frac{9}{TN} \ln(\frac{2 r_{\lambda}}{T} \sum_{t=1}^{T} \frac{1}{\lambda_t})}  + \epsilon \} \leq exp \{- \frac{2TN \epsilon^2}{9}\}
\label{eq:proof_bound_any_lambda_3}
\end{equation}

\noindent
holds for any $\boldsymbol{w} \in \mathcal{F}_{\boldsymbol{\lambda}}$, which is the conclusion of \thmref{thm:bound_any_lambda}, by noting the following facts:

\begin{itemize}
\item Fact 1: if $\boldsymbol{w} \in \mathcal{F}_{\boldsymbol{\lambda}}$, then $\boldsymbol{w} \in \mathcal{F}_{\boldsymbol{\lambda}_{\hat{\boldsymbol{k}}}}$.
\item Fact 2: $\hat{er}_{\boldsymbol{\lambda}} \geq \hat{er}_{\boldsymbol{\lambda}_{\hat{\boldsymbol{k}}}}$ .
\item Fact 3: $\sqrt{2}R(\mathcal{F}_{\boldsymbol{\lambda}}) \geq R(\mathcal{F}_{\boldsymbol{\lambda}_{\hat{\boldsymbol{k}}}})$ .
\item Fact 4: $\epsilon + \sqrt{\frac{9}{TN} \ln(\frac{2 r_{\lambda}}{T} \sum_{t=1}^{T} \frac{1}{\lambda_t})} \geq \epsilon_{\hat{\boldsymbol{k}}}$ .
\end{itemize}

In the following, we give the proof of inequality (\ref{eq:proof_bound_any_lambda_1}), Fact 3 and Fact 4. We omit the proof of Fact 1 and Fact 2, since they are obvious by noticing that $\boldsymbol{\lambda} \succ \boldsymbol{\lambda}_{\hat{\boldsymbol{k}}}$.

\subsubsection{Proof to Inequality (\ref{eq:proof_bound_any_lambda_1})}

According to \thmref{thm:bound_fixed_lambda}, we know that for fixed $r_{\lambda} \geq 1$, $\rho > 0$, $\boldsymbol{\lambda} = [\lambda_1, \cdots, \lambda_T]', \lambda_t \in [1, r_{\lambda}], \forall t = 1,\cdots,T$, and for any $\boldsymbol{w} \in \mathcal{F}_{\boldsymbol{\lambda}}$, $\epsilon > 0$, 

\begin{equation}
P\{er - \hat{er}_{\boldsymbol{\lambda}} > \frac{r_{\lambda}}{\rho} R(\mathcal{F}_{\boldsymbol{\lambda}}) + \epsilon \} \leq exp \{- \frac{2TN \epsilon^2}{9}\}
\label{eq:proof_bound_any_lambda_4}
\end{equation}

\noindent
Given the definition of $\boldsymbol{k} = [k_1, \cdots, k_T]'$, we know that (\ref{eq:proof_bound_any_lambda_4}) holds for all $\{ \epsilon_{\boldsymbol{k}} \}_{k_1, \cdots, k_T}$ and $\{ \boldsymbol{\lambda}_{\boldsymbol{k}} \}_{k_1, \cdots, k_T}$. Therefore, based on the union bound, we have 

\begin{equation}
\begin{aligned}
& P\{ \exists \boldsymbol{k} = [k_1, \cdots, k_T]' : er - \hat{er}_{\boldsymbol{\lambda}_{\boldsymbol{k}}} > \frac{r_{\lambda}}{\rho} R(\mathcal{F}_{\boldsymbol{\lambda}_{\boldsymbol{k}}}) + \epsilon_{\boldsymbol{k}} \} \\
& \leq \sum_{k_1=2}^{r_{\lambda}} \cdots \sum_{k_T=2}^{r_{\lambda}} exp \{- \frac{2TN \epsilon_{\boldsymbol{k}}^2}{9}\} \\
& = \sum_{k_1=2}^{r_{\lambda}} \cdots \sum_{k_T=2}^{r_{\lambda}} exp \left \{ -\frac{2TN}{9} \left (\epsilon + \sqrt{\frac{9\ln \frac{\sum_{t=1}^{T} k_t}{T}}{TN}} \right )^2 \right \} \\
& \leq \sum_{k_1=2}^{r_{\lambda}} \cdots \sum_{k_T=2}^{r_{\lambda}} exp \left \{ -\frac{2TN \epsilon^2}{9} \right \} exp \left \{ -2 \ln \frac{\sum_{t=1}^{T} k_t}{T} \right \} \\
& \leq exp \left \{ -\frac{2TN \epsilon^2}{9} \right \} \sum_{k_1=2}^{r_{\lambda}} \cdots \sum_{k_T=2}^{r_{\lambda}} exp \left \{ - \frac{2}{T} \sum_{t=1}^{T} \ln k_t \right \} \\
& =  exp \left \{ -\frac{2TN \epsilon^2}{9} \right \} \sum_{k_1=2}^{r_{\lambda}} \cdots \sum_{k_T=2}^{r_{\lambda}} exp \left \{ \sum_{t=1}^{T} \ln (k_t)^{- \frac{2}{T}} \right \} \\
& =  exp \left \{ -\frac{2TN \epsilon^2}{9} \right \} \sum_{k_1=2}^{r_{\lambda}} \cdots \sum_{k_T=2}^{r_{\lambda}} \prod_{t=1}^{T} \left (\frac{1}{k_t^2} \right )^{\frac{1}{T} } \\
& \leq exp \left \{ -\frac{2TN \epsilon^2}{9} \right \} \sum_{k_1=2}^{r_{\lambda}} \cdots \sum_{k_T=2}^{r_{\lambda}}  \frac{1}{T} \sum_{t=1}^{T} \left (\frac{1}{k_t^2} \right ) \\
& = exp \left \{ -\frac{2TN \epsilon^2}{9} \right \} \sum_{k=2}^{r_{\lambda}} \frac{1}{k^2} \\
& \leq exp \left \{ -\frac{2TN \epsilon^2}{9} \right \} \sum_{k=2}^{\infty} \frac{1}{k^2} \\
& = exp \left \{ -\frac{2TN \epsilon^2}{9} \right \} (\frac{\pi^2}{6}-1) \\
& \leq exp \left \{ -\frac{2TN \epsilon^2}{9} \right \}
\end{aligned} 
\label{eq:proof_bound_any_lambda_5}
\end{equation}

\subsubsection{Proof to Fact 3}

First, we observe that

\begin{equation}
\lambda_t \leq \lambda_{\hat{k}_t - 1} = \frac{\hat{k}_t}{\hat{k}_t - 1} \lambda_{\hat{k}_t}, \forall t = 1,\cdots,T
\label{eq:proof_bound_any_lambda_6}
\end{equation}

\noindent
Since $2 \leq \hat{k}_t \leq r_{\lambda}$, we know that $\frac{\hat{k}_t}{\hat{k}_t - 1} \leq 2$, which gives $\lambda_t \leq 2 \lambda_{\hat{k}_t}, \forall t = 1,\cdots,T$. Then, based on \thmref{thm:monotone_erc}, we know that $R(\mathcal{F}_{\boldsymbol{\lambda}}) \geq R(\mathcal{F}_{2\boldsymbol{\lambda}_{\hat{\boldsymbol{k}}}})$. Based on the definition of $R(\mathcal{F}_{2\boldsymbol{\lambda}_{\hat{\boldsymbol{k}}}})$:

\begin{equation}
\begin{aligned}
R(\mathcal{F}_{2\boldsymbol{\lambda}_{\hat{\boldsymbol{k}}}}) & = \frac{2}{TN}E[\sup_{\sum_{t=1}^T 2\lambda_{\hat{k}_t}\| \boldsymbol{w}_t \|^2 \leq R } \sum_{t=1}^T \sum_{i=1}^N \sigma_t^i \langle \boldsymbol{w}_t, \phi(\boldsymbol{x}_t^i) \rangle] \\
& = \frac{2}{TN}E[\sup_{\sum_{t=1}^T \lambda_{\hat{k}_t}\| \boldsymbol{v}_t \|^2 \leq R } \sum_{t=1}^T \sum_{i=1}^N \frac{1}{\sqrt{2}}\sigma_t^i \langle \boldsymbol{v}_t, \phi(\boldsymbol{x}_t^i) \rangle] \\
& = \frac{1}{\sqrt{2}} R(\mathcal{F}_{\boldsymbol{\lambda}_{\hat{\boldsymbol{k}}}})
\end{aligned}
\label{eq:proof_bound_any_lambda_7}
\end{equation}

\noindent
Note that the second equality is based on the variable change $\boldsymbol{v}_t \triangleq \sqrt{2} \boldsymbol{w}_t, \forall t = 1,\cdots,T$.

\subsubsection{Proof to Fact 4}
Recall that we defined $\epsilon_{\boldsymbol{k}} \triangleq \epsilon + \sqrt{\frac{9\ln \frac{\sum_{t=1}^{T} k_t}{T}}{TN}}$. Since $\hat{k}_t = \frac{r_{\lambda}}{\lambda_{\hat{k}_t}}$, we know that 

\begin{equation}
\epsilon_{\boldsymbol{k}} \triangleq \epsilon + \sqrt{\frac{9}{TN} \ln \frac{r_{\lambda}}{T} \sum_{t=1}^{T} \frac{1}{\lambda_{\hat{k}_t}}}
\label{eq:proof_bound_any_lambda_8}
\end{equation}

\noindent
As we have shown earlier, $\lambda_t \leq 2 \lambda_{\hat{k}_t}, \forall t = 1,\cdots,T$. Therefore, $\frac{1}{\lambda_{\hat{k}_t}} \leq \frac{2}{\lambda_t}$, which completes the proof.

\subsection{Proof to \thmref{thm:erc_upper_bound}}
\label{sec:proof_erc_upper_bound}
Given the definition of $R(\mathcal{F}_{\boldsymbol{\lambda}})$ in \eref{eq:erc}, by letting $\boldsymbol{v}_t \triangleq \sqrt{\lambda_t} \boldsymbol{w}_t, \forall t = 1, \cdots, T$, we have that 

\begin{equation}
R(\mathcal{F}_{\boldsymbol{\lambda}}) \triangleq \frac{2}{TN}E[\sup_{\boldsymbol{v} \in \mathcal{F}_{\boldsymbol{1}}} \sum_{t=1}^T \sum_{i=1}^N \frac{1}{\sqrt{\lambda_t}} \sigma_t^i \langle \boldsymbol{v}_t, \phi(\boldsymbol{x}_t^i) \rangle]
\label{eq:proof_erc_upper_bound_1}
\end{equation}

\noindent
The proof is completed after using the Cauchy-Schwarz inequality.

\subsection{Proof to \propref{prop:model_equivalent}}
\label{sec:proof_model_equivalent}
First of all, based on \propref{prop:kloft2011}, for any fixed $C > 0$, $s > 0$ and $r_{\lambda} > 0$, there exist $R > 0$ and $a > 0$ such that \pref{eq:model} and the following optimization problem are equivalent

\begin{equation}
\begin{aligned}
\min_{\boldsymbol{w}, \boldsymbol{\lambda}} \; & \sum_{t=1}^{T} \lambda_t (\| \boldsymbol{w}_t \|^2 + C \sum_{i=1}^{N} l(y_t^i\langle \boldsymbol{w}_t, \phi(\boldsymbol{x}_t^i) \rangle)) \\ 
\textit{s.t.} \; & \boldsymbol{w}_t \in \mathcal{H}_{\boldsymbol{\theta}}, \forall t = 1, \cdots, T,\\
& \boldsymbol{\theta} \in \Omega(\boldsymbol{\theta}), \; \boldsymbol{1} \prec \boldsymbol{\lambda} \prec r_{\lambda}\boldsymbol{1}, \sqrt{\sum_{t=1}^{T} \frac{1}{\lambda_t}} \leq \sqrt{a}.
\end{aligned}
\label{eq:proof_model_equivalent_1}
\end{equation}

\noindent
where the constraint $\sqrt{\sum_{t=1}^{T} \frac{1}{\lambda_t}} \leq \sqrt{a}$ is equivalent to $\sum_{t=1}^{T} \frac{1}{\lambda_t} \leq a$. Then, since $\boldsymbol{w}_t \in \mathcal{H}_{\boldsymbol{\theta}}$, there must exist $\boldsymbol{w}_t^m \in \mathcal{H}_m, \forall t = 1,\cdots,T, m = 1,\cdots,M$, such that $\boldsymbol{w}_t = [\sqrt{\theta_1}{\boldsymbol{w}_t^1}', \cdots, \sqrt{\theta_M}{\boldsymbol{w}_t^M}']'$. Similarly, $\phi(\boldsymbol{x}) = [\sqrt{\theta_1}\phi_1(\boldsymbol{x})', \cdots, \sqrt{\theta_M}\phi_M(\boldsymbol{x})']'$. We complete the proof by substituting these two equalities into \pref{eq:proof_model_equivalent_1} and letting $\boldsymbol{v}_t^m \triangleq \theta_m \boldsymbol{w}_t^m, \forall t = 1,\cdots,T, m = 1,\cdots,M$.

\subsection{Proof to \thmref{thm:erc_upper_bound_mkl}}
\label{sec:proof_erc_upper_bound_mkl}
Starting with the maximization problem in \eref{eq:proof_erc_upper_bound_1}, it is not hard to see that it is equivalent to

\begin{equation}
\begin{aligned}
\sup_{\boldsymbol{\alpha}, \boldsymbol{\theta}} & \sum_{t=1}^T  \frac{1}{\sqrt{\lambda_t}} {\boldsymbol{\sigma}_t}' \boldsymbol{K}_t \boldsymbol{\alpha}_t \\
\textit{s.t.} & \; \sum_{t=1}^{T} {\boldsymbol{\alpha}_t}' \boldsymbol{K}_t \boldsymbol{\alpha}_t \leq R \\
& \boldsymbol{\theta} \succeq \boldsymbol{0}, \| \boldsymbol{\theta} \|_p \leq 1.
\end{aligned}
\label{eq:proof_erc_upper_bound_mkl_1}
\end{equation}

\noindent
where $\boldsymbol{K}_t = \sum_{m=1}^{M} \theta_m \boldsymbol{K}_t^m$. Let $\tilde{\boldsymbol{\sigma}}_t \triangleq \boldsymbol{\sigma}_t / \sqrt{\lambda_t}$, $\tilde{\boldsymbol{\sigma}} \triangleq [{\tilde{\boldsymbol{\sigma}}_1}', \cdots, {\tilde{\boldsymbol{\sigma}}_T}']'$, $\boldsymbol{\alpha} \triangleq [{\boldsymbol{\alpha}_1}', \cdots, {\boldsymbol{\alpha}_T}']'$, and $\boldsymbol{K}$ be the block diagonal matrix, with the diagonal blocks be the $\boldsymbol{K}_t$'s, \pref{eq:proof_erc_upper_bound_mkl_1} becomes

\begin{equation}
\begin{aligned}
\sup_{\boldsymbol{\alpha}, \boldsymbol{\theta}} &\; \tilde{\boldsymbol{\sigma}}' \boldsymbol{K \alpha} \\
\textit{s.t.} & \; \boldsymbol{\alpha}' \boldsymbol{K \alpha} \leq R \\
& \boldsymbol{\theta} \succeq \boldsymbol{0}, \| \boldsymbol{\theta} \|_p \leq 1.
\end{aligned}
\label{eq:proof_erc_upper_bound_mkl_2}
\end{equation}

\noindent
Optimizing with respect to $\boldsymbol{\alpha}$ yields the closed-form solution: $\boldsymbol{\alpha}^* = \sqrt{\frac{R}{\tilde{\boldsymbol{\sigma}}' \boldsymbol{K} \tilde{\boldsymbol{\sigma}}}} \boldsymbol{K} \tilde{\boldsymbol{\sigma}}$, and therefore we have 

\begin{equation}
R(\mathcal{F}_{\boldsymbol{\lambda}}) \leq \frac{2}{TN} E \left [\sup_{\boldsymbol{\theta} \succeq \boldsymbol{0}, \| \boldsymbol{\theta} \|_p \leq 1} \; \sqrt{R \sum_{t=1}^{T} \frac{1}{\lambda_t} {\boldsymbol{\sigma}_t}' \boldsymbol{K}_t \boldsymbol{\sigma}_t } \; \right ]
\label{eq:proof_erc_upper_bound_mkl_3}
\end{equation}

\noindent
Let $\boldsymbol{u}_t \triangleq [u_t^1, \cdots, u_t^M]'$  with  ${u_t^m \triangleq \boldsymbol{\sigma}_t}' \boldsymbol{K}_t^m \boldsymbol{\sigma}_t, m = 1,\cdots,M$, we have 

\begin{equation}
\begin{aligned}
R(\mathcal{F}_{\boldsymbol{\lambda}}) & \leq \frac{2\sqrt{R}}{TN} E \left [ \sup_{\boldsymbol{\theta} \succeq \boldsymbol{0}, \| \boldsymbol{\theta} \|_p \leq 1} \; \sqrt{\sum_{t=1}^{T} \frac{1}{\lambda_t} \boldsymbol{\theta}' \boldsymbol{u}_t } \; \right ] \\
& \leq \frac{2\sqrt{R}}{TN} E \left [\sup_{\boldsymbol{\theta} \succeq \boldsymbol{0}, \| \boldsymbol{\theta} \|_p \leq 1} \; \sqrt{\sum_{t=1}^{T} \frac{1}{\lambda_t} \| \boldsymbol{\theta} \|_p \| \boldsymbol{u}_t \|_{p^*} } \; \right ] \\
& = \frac{2\sqrt{R}}{TN} E \left [\sqrt{\sup_{\boldsymbol{\theta} \succeq \boldsymbol{0}, \| \boldsymbol{\theta} \|_p \leq 1} \;\sum_{t=1}^{T} \frac{1}{\lambda_t} \| \boldsymbol{\theta} \|_p \| \boldsymbol{u}_t \|_{p^*} } \; \right ] \\
& = \frac{2\sqrt{R}}{TN} E \left [\sqrt{\sum_{t=1}^{T} \frac{1}{\lambda_t} \| \boldsymbol{u}_t \|_{p^*} } \; \right ] \\
& \leq  \frac{2\sqrt{R}}{TN} \sqrt{\sum_{t=1}^{T} \frac{1}{\lambda_t} \left ( \sum_{m=1}^{M} E (u_t^m)^{p^*} \right )^{\frac{1}{p^*}} } \\
& =  \frac{2\sqrt{R}}{TN} \sqrt{\sum_{t=1}^{T} \frac{1}{\lambda_t} \left ( \sum_{m=1}^{M} E (\| \sum_{i=1}^{N} \sigma_t^i \phi_m(\boldsymbol{x}_t^i) \|)^{2p^*} \right )^{\frac{1}{p^*}} }
\end{aligned} 
\label{eq:proof_erc_upper_bound_mkl_4}
\end{equation}

\noindent
where the last inequality is due to Jensen's Inequality. Finally, the proof is completed by utilizing the following inequality, which holds for any $\phi : \mathcal{X} \mapsto \mathcal{H}$ and $p \geq 1$:

\begin{equation}
E_{\sigma} \| \sum_{i=1}^n \sigma_i \phi(\boldsymbol{x}_i) \|^p \leq  (p \sum_{i=1}^n \| \phi(\boldsymbol{x}_i) \|^2)^{\frac{p}{2}}
\label{eq:proof_erc_upper_bound_mkl_5}
\end{equation}

\end{document}